\documentclass[11pt]{article}

\usepackage[final]{acl}

\usepackage{times}
\usepackage{latexsym}

\usepackage[T1]{fontenc}

\usepackage[utf8]{inputenc}

\usepackage{microtype}

\usepackage{inconsolata}

\usepackage{graphicx}
\usepackage{amsmath}
\usepackage{amssymb}
\usepackage{amsfonts}
\usepackage{booktabs}
\usepackage{tabularx}
\usepackage{enumitem}
\title{From Isolation to Alignment: \\ Unified LoRA for Efficient Multi-Task Learning}

\author{
        Jinda Liu$^{1,2}$ \quad  Yi Chang$^{1,3,4}$ \quad Yuan Wu$^{1}$\footnotemark[1] \\
        $^{1}$School of Artificial Intelligence, Jilin University \\
        $^{2}$Southern University of Science and Technology \\
        $^{3}$International Center of Future Science, Jilin University\\
        $^{4}$Engineering Research Center of Knowledge-Driven Human-Machine Intelligence, MOE, China \\
        liujd9922@mails.jlu.edu.cn, 12632277@mail.sustech.edu.cn \\
        \{yichang, yuanwu\}@jlu.edu.cn \\
}

\begin{document}
\maketitle
\renewcommand{\thefootnote}{\fnsymbol{footnote}}
\footnotetext[1]{Corresponding author}

\begin{abstract}
Parameter-Efficient Fine-Tuning (PEFT) is essential for adapting Large Language Models (LLMs) to multi-task scenarios. A prevailing trend in this field involves complex LoRA variants with multiple adapters or heads, which rely on the premise that architectural isolation of task-specific knowledge is necessary. However, this design often introduces dynamic routing, preventing weight merging and causing significant inference latency. In this work, we present a direct challenge to this paradigm. We first reveal a paradox where a simplified, router-free multi-head model with high inter-head redundancy outperforms complex, diversity-driven baselines. Furthermore, we demonstrate that a unified, single-adapter LoRA with increased rank achieves highly competitive performance, questioning the necessity of multi-component structures. Based on these findings, we propose Align-LoRA, a unified and efficient framework that shifts the focus from architectural isolation to representation alignment. Align-LoRA incorporates an explicit alignment loss to encourage the learning of task-shared representations within a shared latent space. Crucially, our method maintains the standard LoRA architecture, ensuring zero inference latency via weight merging. Theoretical analysis and extensive experiments confirm that Align-LoRA significantly surpasses prevailing approaches, establishing a simpler, more effective, and production-friendly paradigm for multi-task PEFT.
The code is available at \href{https://github.com/jinda-liu/Align-LoRA}{Align-LoRA}. 
\end{abstract}

\section{Introduction}
\label{introduction}
In recent years, large language models (LLMs) have demonstrated unprecedented performance across a wide range of natural language processing (NLP) tasks \citep{brown2020language,zhao2026survey,chang2024survey}. Despite their strong generalization abilities, LLMs often require further adaptation to align with domain-specific requirements or to incorporate updated knowledge~\citep{yang2025mtl,xin2024beyond}. Supervised fine-tuning (SFT) plays a critical role in this process, but full parameter fine-tuning (FFT), which updates all model parameters, poses significant challenges in terms of computational and memory costs~\citep{mao2025survey}.

To address these demands, parameter-efficient fine-tuning (PEFT) methods have been proposed to adapt LLMs by updating only a small subset of parameters~\citep{hu2022lora,chang2026ba}. Among these, Low-Rank Adaptation (LoRA)~\citep{hu2022lora} has become a widely adopted approach. It approximates the full-rank weight update matrix by decomposing it into two low-rank matrices: a down-projection matrix $\mathbf{A}$ and an up-projection matrix $\mathbf{B}$. In practice, adapting LLMs often involves data from multiple domains or tasks, naturally aligning with the multi-task learning paradigm.

\begin{figure}[t]
\includegraphics[width=0.99\columnwidth]{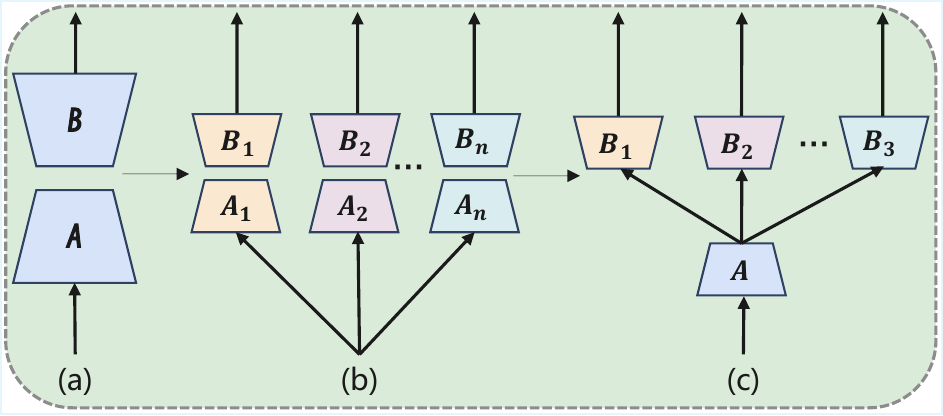} 
\centering
\caption{A comparison of LoRA architectural paradigms: (a) Vanilla LoRA; (b) the Multi-Adapter framework; and (c) the Multi-Head framework. The categorization of the multi-component architectures is adapted from R-LoRA~\citep{liu-rlora}. A common feature of these multi-component designs is the inclusion of a dynamic routing mechanism.}
\label{architecture}
\end{figure}

Consequently, this has motivated the development of LoRA variants specifically designed for MTL. An early approach is the Multi-Adapter architecture, which employs multiple, distinct pairs of down-projection ($\mathbf{A}$) and up-projection ($\mathbf{B}$) matrices for different tasks~\citep{wang2023multilora}. To improve parameter efficiency, the Multi-Head architecture was introduced, typically sharing a single $\mathbf{A}$ matrix while maintaining multiple task-specific head matrices ($\mathbf{B}$)~\citep{tian2024hydralora}. Furthermore, many of these multi-component architectures employ a routing mechanism, inspired by the Mixture-of-Experts (MoE) framework, to dynamically select or weigh the outputs of different adapters for a given input. Recent prevalent methods like R-LoRA~\citep{liu-rlora} further refine this by explicitly encouraging diversity among heads to mitigate redundancy. Figure~\ref{architecture} illustrates these paradigms. \textbf{Despite architectural differences, these methods are all built on a common premise: that effective multi-task adaptation requires architectural isolation of task-specific knowledge.}

However, architectural isolation has a practical cost: specialized components cannot be merged into the backbone, incurring non-negligible inference latency. This trade-off motivates our work. We first reveal a paradoxical finding: M-LoRA, a simplified model that removes the dynamic router, surpasses its more complex counterparts despite exhibiting higher inter-head similarity. This directly challenges the assumption that component diversity is beneficial, leading to a fundamental question: \textbf{Is the multi-component structure truly necessary for effective multi-task adaptation?}

Collectively, these findings point to a new hypothesis: \textbf{learning task-shared representations provides a promising alternative to the architectural isolation of task-specific features.}
To directly validate this hypothesis and operationalize this principle, we propose Align-LoRA. This method enhances a standard LoRA by augmenting its training objective with a component based on the Kullback-Leibler (KL) Divergence ~\citep{kullback1951information}, which encourages the alignment of task representations in the shared low-rank space without adding parameters or inference overhead.

Our key contributions are three-fold:
\begin{itemize}
    \item \textbf{Challenging Architectural Isolation:} We demonstrate that simple baselines, specifically a redundant \textbf{M-LoRA} and a standard LoRA with \textbf{increased rank}, consistently outperform complex multi-component variants. This finding critically questions the necessity of architectural isolation and the additional inference latency it introduces.

    \item \textbf{The Align-LoRA Framework:} We challenge the dominant paradigm that multi-task LoRA requires structural isolation of task-specific knowledge. Instead, we propose and validate that explicitly aligning shared representations in the LoRA latent space is a more effective alternative---realized through \textbf{Align-LoRA}, which adds only a representation-level alignment loss to standard LoRA while preserving zero inference latency.

    \item \textbf{Rigorous Validation:} We provide a theoretical analysis of Align-LoRA's effectiveness by deriving a tighter generalization bound. Extensive experiments across diverse models, scales, and task benchmarks validate the practical effectiveness of this principle.
\end{itemize}

\begin{figure*}[t]
    \centering
    \begin{minipage}[t]{0.63\textwidth}
        \vspace{0pt}
        \centering
        \includegraphics[width=\linewidth]{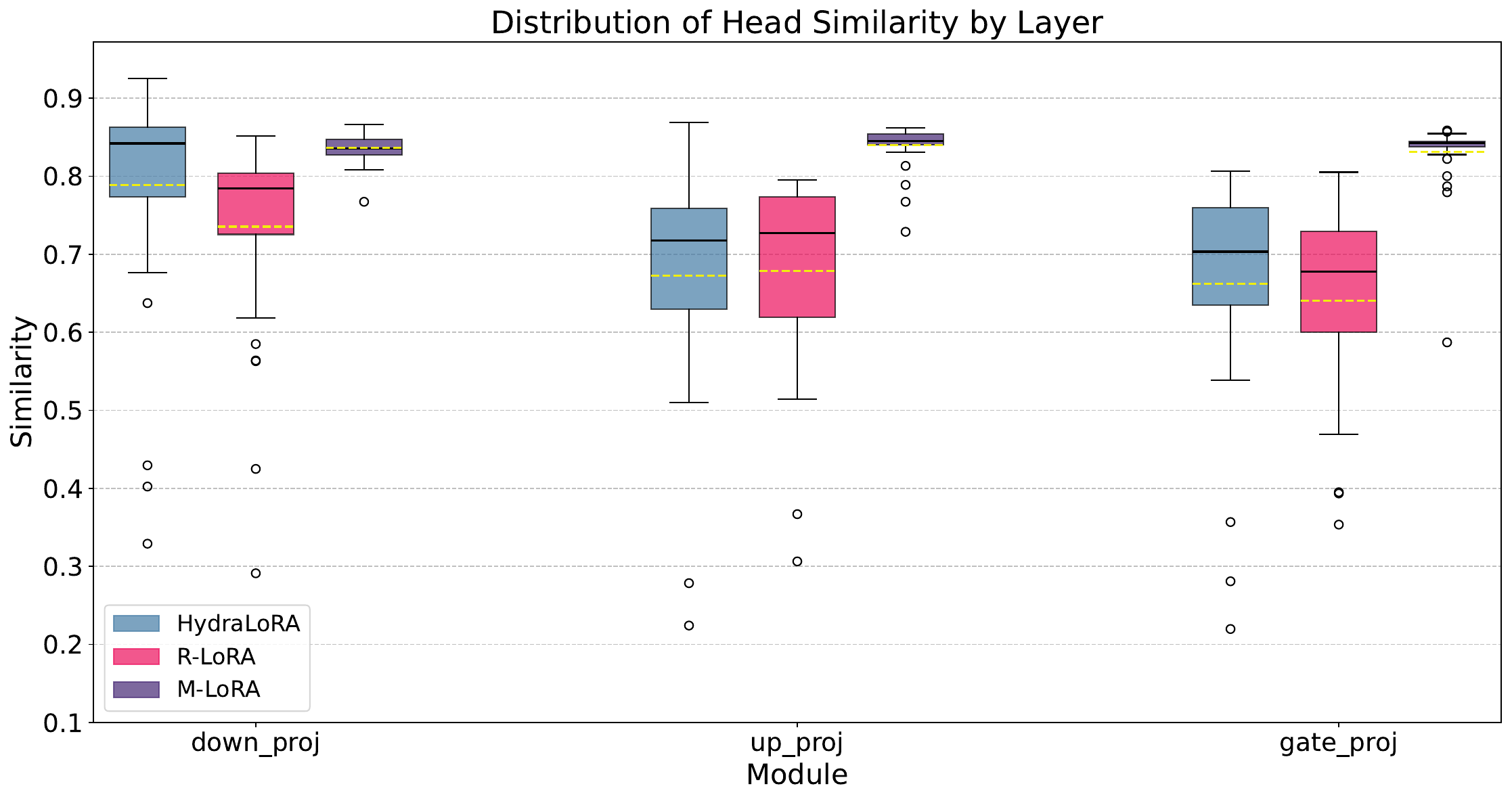}
    \end{minipage}
    \hfill
    \begin{minipage}[t]{0.36\textwidth}
        \vspace{0pt}
        \centering
        \includegraphics[width=\linewidth]{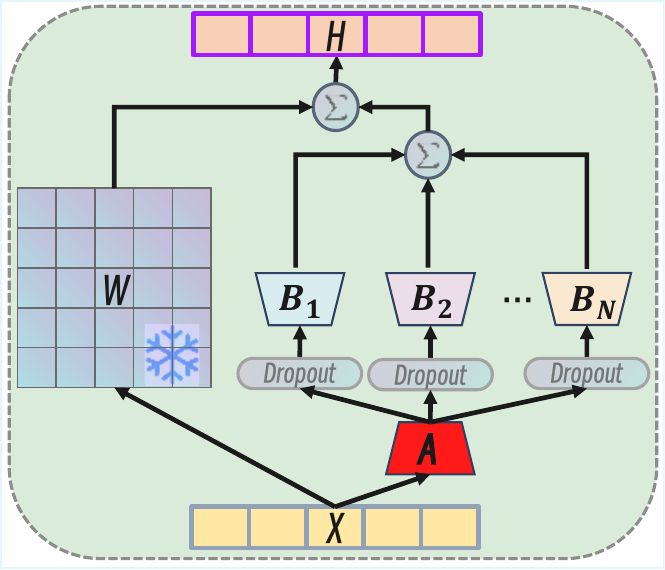}
    \end{minipage}
    
    \vspace{-0.5em} 
    
    \begin{minipage}[t]{0.62\textwidth}
        \caption{Distribution of inter-head cosine similarity across different model modules. Within each module, the box plots from left to right correspond to HydraLoRA, R-LoRA, and M-LoRA, respectively. The solid line within each box indicates the median similarity, while the dashed line represents the mean.}
        \label{cosin}
    \end{minipage}
    \hfill
    \begin{minipage}[t]{0.35\textwidth}
        \caption{The framework of M-LoRA. Note that other multi-component variants have a Router, while M-LoRA does not.}
        \label{fig: M-LoRA}
    \end{minipage}
\end{figure*}

\begin{table*}
  \centering
  \caption{Comparative study of several multi-head LoRA variants across five tasks.}
  \begin{tabular}{l | ccccccc}
    \hline
    \textbf{Schemes} & \textbf{QNLI} & \textbf{PiQA} & \textbf{Winogrande} & \textbf{ARC} & \textbf{GSM8K} & \textbf{Avg} & \textbf{\%Para}\\
    \hline
    HydraLoRA    & 81.91 & 84.21 & 70.92 & 87.21 & 45.95 & 74.04 & 0.45\\
     \quad w/o Router    & 81.33 & 83.51 & 70.14 & 86.46 & 45.35 & 73.58 & 0.41\\
    R-LoRA   & 82.03 & 85.55 & 71.84 & 87.69 & 46.25 & 74.67  & 0.45 \\
    M-LoRA   & \textbf{82.52} & \textbf{86.76} & \textbf{72.95} & \textbf{88.15} & \textbf{46.85} & \textbf{75.45} & \textbf{0.41} \\
    \hline
  \end{tabular}
  \label{table: observation}
\end{table*}

\begin{table*}
  \caption{Comparison of different training schemes on LLaMA2. All variants use a LoRA rank of 8, except $\text{LoRA}^{\dagger}$(rank=30, adjusted to match multi-head variants’ trainable parameter count). * indicates results from \cite{tian2024hydralora}.}
  \centering
  \renewcommand{\arraystretch}{0.9}
  \setlength{\tabcolsep}{4pt}
  \begin{tabular}{l | ccccccccc}
    \hline
    \textbf{Metrics} & \textbf{Base} & \textbf{LoRA} & \textbf{LoRAHub*}& \textbf{LoRA MoE*} & \textbf{HydraLoRA} & \textbf{R-LoRA} & \textbf{$\text{LoRA}^{\dagger}$} & \textbf{M-LoRA}\\
    \hline
    7B     & 31.61 & 37.05 & 39.70 & 40.30 & 41.46 & \underline{42.24} & 42.21 & \textbf{42.83} \\
    13B & 38.42 & 40.73 & 41.90 & 43.70 & 44.31 & 44.96 & \underline{45.02} & \textbf{46.16}\\
    \% Param & - & 0.06 & 1.24 & 2.98 & 0.34 & 0.34 & 0.34 & 0.32\\
    \hline
  \end{tabular}
  \label{llama-flan}
\end{table*}

\begin{table*}
  \centering
  \renewcommand{\arraystretch}{0.9}
  \caption{Comparison of different training schemes on Qwen2.5. The superscript in "LoRA" (e.g., \textsuperscript{4}, \textsuperscript{8}, etc.) indicates the rank value used for each variant.}
  \begin{tabular}{l | cccccccc}
    \hline
    \textbf{Metrics} & \textbf{Base}& \textbf{$\text{LoRA}^{4}$} & \textbf{$\text{LoRA}^{8}$} & \textbf{$\text{LoRA}^{9}$} & \textbf{$\text{LoRA}^{10}$} & \textbf{HydraLoRA} & \textbf{R-LoRA} & \textbf{M-LoRA}\\
    \hline
    7B    & 39.82 & 43.21 & 46.66 & 48.18 & \underline{49.51} & 49.12 &  \underline{49.51} &  \textbf{49.74}\\
    14B   & 45.33  & 48.18 & 51.82 & 52.74 & \textbf{54.23} & 53.76 &  54.08 &  \underline{54.18}\\
     Rank & - & 4 & 8 & 9 & 10 & 4 & 4 & 4\\
    \% Param & - & 0.10 & 0.20 & 0.22 & 0.25 & 0.25 & 0.25 & 0.22\\

    \hline
  \end{tabular}
  \label{qwen-flan}
\end{table*}

\section{Observations}
\label{observation}
In this section, we critically examine the prevailing assumption that component diversity are essential for effective multi-task adaptation with LoRA. By questioning the fundamental necessity of the prevalent multi-head paradigm, our investigation leads to a new hypothesis centered on the pivotal role of shared knowledge.

\subsection{M-LoRA: A Simplified Variant}

R-LoRA~\citep{liu-rlora} encourages head diversity for task-specific knowledge. To test this hypothesis, we propose \textbf{M-LoRA}, an ablation that removes R-LoRA's dynamic router while preserving multi-head randomization and Dropout. M-LoRA aggregates head outputs by simple summation, directly testing whether explicit diversification is necessary. The framework is illustrated in Appendix~\ref{M-LoRA}.


\subsection{The Paradox of Diversity: Less is More}
\label{sec:paradox_of_diversity}
We fine-tune the Qwen2.5-3B~\citep{qwen2.5} model using HydraLoRA, R-LoRA, and M-LoRA on a benchmark comprising five distinct tasks: QNLI~\citep{wang2018glue}, PiQA~\citep{bisk2020piqa}, Winogrande~\citep{sakaguchi2021winogrande}, ARC (easy \& challenge)~\citep{allenai:arc}, and GSM8K~\citep{cobbe2021training}. To quantify inter-head similarity, we compute a matrix of pairwise cosine similarities between all flattened head vectors ($\mathbf{B}_i$). The final metric is the mean of this matrix's off-diagonal values. All Experimental details in this work, including implementation specifics, hyperparameter settings, dataset descriptions, baseline configurations, and other relevant information, are documented in the Appendix~\ref{dataset}.

Our findings reveal a paradox regarding the role of head diversity in multi-task adaptation. 
Figure~\ref{cosin}, which plots the inter-head cosine similarity, shows that R-LoRA successfully achieves its design goal of maximizing diversity, exhibiting the lowest similarity. 
In stark contrast, M-LoRA, which lacks any diversity-enforcing mechanism, displays the opposite effect, yielding a high degree of head redundancy with similarity medians consistently exceeding 0.85.
Paradoxically, as demonstrated in Table~\ref{table: observation}, this high-redundancy model achieves superior multi-task performance. 
Despite its architectural simplicity, M-LoRA consistently and significantly outperforms the more complex HydraLoRA and R-LoRA. 
This outcome presents a fundamental contradiction to the philosophy of prior work: the architectural configuration that seemingly violates the principle of head diversity actually enhances multi-task generalization.

A natural concern is that this advantage may stem from head count coinciding with task count ($N{=}5$), implicitly allowing each head to specialize. To test this, we vary $N$ while keeping five tasks fixed (see Appendix~\ref{sec:head_ablation}). Even when heads must be shared across tasks ($N{=}2$ or $N{=}3$), M-LoRA remains competitive (74.14 and 74.62 vs.\ 75.45 at $N{=}5$), confirming that the router-free design handles task sharing without explicit routing mechanisms. The slight advantage at $N{=}5$ reflects the natural correspondence between heads and tasks, not a dependency on this configuration.

Extended training (1--3 epochs) in Appendix~\ref{sec:extended_train} shows that while R-LoRA's router theoretically subsumes M-LoRA, the additional parameters create a harder optimization landscape; M-LoRA's advantage not only persists but widens ($+0.78 \to +1.23$pp, see Appendix~\ref{sec:extended_train}), ruling out convergence artifacts.

\subsection{From Observation to Hypothesis}
\label{sec:task_shared_vs_specific}
This section explains M-LoRA's effectiveness. Its strong multi-task generalization challenges conventional assumptions and offers a new perspective on multi-task adaptation in LoRA.
Improving multi-task learning has followed two distinct paths: isolating \textbf{task-specific} knowledge to mitigate interference, or enhancing \textbf{task-shared} knowledge to improve generalization. While recent methods like LoRA MoE~\citep{dou-etal-2024-loramoe} and R-LoRA~\citep{liu-rlora} focus on isolating task-specific knowledge, \textbf{the alternative path of actively enhancing task-shared knowledge within the LoRA framework remains unexplored.}

\textbf{M-LoRA} challenges this specialization-focused paradigm. We hypothesize that high similarity is not a sign of failed specialization, but a feature derived from the architecture's implicit regularization. The key mechanism is the interplay between removing the router and retaining multi-head dropout.
In models like R-LoRA, heads act as competing "specialists" via a dynamic router, which can lead to redundancy or load imbalance. In contrast, by replacing the router with simple averaging, M-LoRA compels the multiple $\mathbf{B}$ heads to form a collaborative ensemble.
As illustrated in Figure~\ref{fig: M-LoRA}, retaining the multi-head dropout forces each head to learn from different input perspectives. By requiring all heads to contribute via summation, they are compelled to converge on a robust, task-general representation.

To validate this mechanism, we performed an ablation on a non-dropout multi-head variant, HydraLoRA. As shown in Table~\ref{table: observation}, removing the router from HydraLoRA (`w/o Router`) causes its performance to drop, while M-LoRA's collaborative ensemble achieves the highest performance. To provide a rigorous causal proof, we conduct a full $2\times 2$ factorial analysis (Router $\times$ Dropout) detailed in Appendix~\ref{sec:causal_attribution}. The empirical results show that removing dropout leads to a severe performance degradation ($-1.63$pp), an effect over $3\times$ larger than that of router removal ($-0.46$pp). This quantitative attribution firmly establishes that dropout-based collaborative ensembling—rather than the mere absence of routing—is the primary driver of M-LoRA's advantage, transforming isolated "specialists" into effective "collaborators" and significantly enhancing task-general learning.

The success of M-LoRA suggests a revised viewpoint: \textbf{learning task-shared representations provides a promising path towards multi-task learning, offering a powerful alternative to the architectural isolation of task-specific features.}
\section{Increasing Rank: A Unified Adapter}
\label{sec: increasing rank}
Given M-LoRA’s strong performance with highly redundant heads, which largely learn shared knowledge, this section explores a critical question: Is the multi-head architecture itself truly necessary for multi-task generalization, or does it merely serve as a means to increase total trainable parameters rather than offering genuine benefits?

To test this, we design a straightforward yet powerful experiment. We abandon the multi-component structure entirely and instead use a standard, single-adapter LoRA. We reallocate the entire parameter budget of the complex variants into this single adapter by simply increasing its rank, $r$. Following the experimental setup of HydraLoRA~\citep{tian2024hydralora}, we conduct fine-tuning on a curated subset of the Flanv2 dataset~\citep{liu2022few}. This training data is sampled from dozens of individual datasets and organized into ten distinct task categories, providing comprehensive training across both NLU and NLG capabilities. We then evaluate the models' multi-task generalization on the challenging BBH(3-shot) benchmark~\citep{suzgun-etal-2023-challenging}, which is designed to test generalization.

The results, presented in Table~\ref{llama-flan} and Table~\ref{qwen-flan}, reveal a clear trend. Across different base models, the performance of a standard LoRA adapter consistently improves with its rank. Crucially, when its rank is scaled to a comparable parameter count, a simple, single-adapter LoRA achieves performance that is competitive with sophisticated multi-component architectures such as LoRA-Hub~\citep{huang2024lorahub}, LoRA MoE~\citep{dou-etal-2024-loramoe}, HydraLoRA~\citep{tian2024hydralora}, and R-LoRA~\citep{liu-rlora}.

This finding provides compelling evidence that \textbf{the architectural complexity of multi-adapter designs may not be a prerequisite for achieving strong multi-task generalization.} Our results indicate that a simple, unified adapter with sufficient capacity delivers comparable performance. This challenges not only the trend toward elaborate structures but also the underlying strategy of isolating task-specific features, suggesting it is a less effective path to generalization than previously assumed and that the research focus on specialized components may warrant reconsideration.

\section{Beyond Rank: Align-LoRA}
\label{beyond rank}
Our investigation in the preceding sections has led to two critical conclusions. First, based on our analysis in Section~\ref{observation}, we formed a guiding hypothesis: learning task-general, shared knowledge may be more critical than enforcing task-specific separation. Second, our findings demonstrate that the architectural complexity introduced by multi-component designs is unnecessary for achieving strong multi-task generalization. This calls into question the prevailing assumption that specialized structures are a prerequisite for effective multi-task learning.

These conclusions motivate a shift from structural isolation toward a more principled approach. By adopting the standard, high-rank LoRA as an efficient baseline, we move beyond merely increasing rank to address a fundamental challenge: \textbf{how to explicitly validate and enhance the learning of shared representations within a unified adapter.} To this end, we introduce \textbf{Align-LoRA}, a novel framework that bridges our theoretical hypothesis with a concrete alignment mechanism.

\begin{table*}
  \centering
  \renewcommand{\arraystretch}{0.9}
  \setlength{\tabcolsep}{4pt}
  \caption{Multi-task generalization performance on the BBH benchmark. Align-LoRA (A-LoRA), particularly A-LoRA-K, demonstrates a clear advantage over the other variants across model families and scales.}
  \begin{tabular}{l | ccccccccc}
    \hline
    \textbf{Metrics} & \textbf{LoRA} & \textbf{LoRAMoE} & \textbf{HydraLoRA} & \textbf{R-LoRA} & \textbf{M-LoRA} & \textbf{A-LoRA-M} & \textbf{A-LoRA-K}\\
    \hline
    Qwen2.5-7B     & 48.36 & 47.18 & 47.38 & 48.32 &  48.44 &  47.53 & \textbf{50.28} \\
    LLaMA3-8B     & 44.89 & 44.18 & 44.03 & 45.01 & 45.35 &  45.42 & \textbf{48.84} \\
    Qwen2.5-14B    & 52.93 & 50.74 & 51.92 & 52.21 &  53.78 &  52.24 & \textbf{55.11} \\
     Rank & 10 & 4 & 4 & 4 & 4 & 8 & 8 \\
    \% Param & 0.25 & 0.38 & 0.25 & 0.25  & 0.22 & \textbf{0.20} & \textbf{0.20}\\
    \hline
  \end{tabular}
  \label{align multi-task}
\end{table*}

\begin{table*}[ht]
  \centering
  \renewcommand{\arraystretch}{0.9}
  \caption{Performance comparison on an 8-task multi-task reasoning benchmark, evaluated on the Qwen2.5-3B and Qwen2.5-7B models. Baselines include the original LoRA and several multi-component variants. Our proposed Align-LoRA (A-LoRA) consistently achieves significantly superior performance, demonstrating its strong multi-task generalization capabilities.}
  \begin{tabular}{l c c c c c c c c c c}
    \hline
    \textbf{Schemes} & \textbf{Task1} & \textbf{2} & \textbf{3} & \textbf{4}  & \textbf{5} & \textbf{6} & \textbf{7} & \textbf{8}& \textbf{Avg} & \textbf{\%Par} \\
    \midrule
    \multicolumn{6}{l}{\textit{Qwen2.5-3B}}\\
    \midrule
    LoRA  & 86.31 & 56.42 & 84.65 & 72.76 & 91.37 & 87.91 & 87.60 & 44.80 & 76.48 & 0.45\\
    LoRAMoE  & 87.41 & 58.21 & 85.64 & 73.37 & 92.18 & 87.40 & 87.35 & 44.80 & 77.05 & 0.68\\
    HydraLoRA & 86.58 & 56.42 & 85.00 & 73.36 & 92.18 & 87.33 & 88.38 & 45.15 & 76.80 & 0.45\\
    R-LoRA  & 87.12 & 57.95 & 88.13 & 73.89 & 94.71 & 88.25 & 88.26 & 45.60 & 77.99 & 0.45\\
    M-LoRA  & 88.02 & 57.95 & 88.87 & 74.21 & 94.71 & 88.91 & 89.07 & 46.35 & 78.51  & 0.42\\
    A-LoRA-M  & 87.94 & 58.03 & 88.87 & 74.12 & 94.51 & 88.85 & 88.61 & 45.88 & 78.35   & 0.42\\
    A-LoRA-K  & \textbf{89.25} & \textbf{59.88} & \textbf{90.35} & \textbf{75.41} & \textbf{95.33} & \textbf{89.55} & \textbf{91.95} & \textbf{48.75} & \textbf{80.06} & \textbf{0.42}\\
    \midrule
    \multicolumn{6}{l}{\textit{Qwen2.5-7B}}\\
    \midrule

    LoRA & 88.41 & 60.78 & 88.42 & 81.58 & 93.52 & 91.20 & 91.79 & 48.15 & 80.48 & 0.25 \\
    LoRAMoE & 89.52 & 61.44 & 88.86 & 82.94 & 92.87 & 91.54 & 91.89 & 48.72 & 80.97 & 0.38 \\
    HydraLoRA & 88.66 & 61.23 & 89.55 & 81.72 & 93.57 & 91.67 & 91.74 & 48.70 & 80.86 & 0.25 \\
    R-LoRA & 89.80 & 62.51 & 89.36 & 83.78 & 95.12 & 91.02 & 92.17 & 50.15 & 81.74 & 0.25 \\
    M-LoRA & 91.35 & 62.51 & 91.98 & 84.70 & 95.93 & 91.02 & 91.97 & 50.20 & 82.46  & 0.22 \\
    A-LoRA-M & 90.86 & 62.45 & 91.68 & 84.59 & 95.93 & 90.74 & 91.75 & 50.45 & 82.31  & 0.20 \\
    A-LoRA-K & \textbf{92.23} & \textbf{64.85} & \textbf{92.89} & \textbf{85.73} & \textbf{95.93} & \textbf{93.35} & \textbf{92.93} & \textbf{53.66} & \textbf{83.95} & \textbf{0.20} \\

    \hline
  \end{tabular}
  \label{multi-task adapt}
\end{table*}

\subsection{Methodology}
To enhance multi-task generalization, we introduce \textbf{Align-LoRA}, a method that encourages the model to learn task-shared features. Align-LoRA introduces an alignment loss, $\mathcal{L}_{\text{align}}$, to explicitly minimize the statistical distance between the low-dimensional representations generated by the shared LoRA down-projection matrix, $\mathbf{A}$. \textbf{To the best of our knowledge, this is the first work to systematically apply statistical distance metrics for this purpose within the multi-task LoRA framework}, drawing inspiration from their foundational use in domain adaptation~\citep{transferpan2010domain}.

Our primary approach instantiates $\mathcal{L}_{\text{align}}$ using the \textbf{Kullback-Leibler (KL) divergence}~\citep{kullback1951information}, a classic statistical method for measuring the distance between distributions. However, we hypothesize that the principle of aligning representations is broadly applicable and not contingent on a single metric. To validate this, we introduce a variant that employs the multi-kernel \textbf{Maximum Mean Discrepancy (MK-MMD)}~\citep{MK-MMDgretton2012optimal}. The strong performance of both instantiations, as we will demonstrate, validates our core thesis: that explicitly aligning the low-dimensional representations of different tasks is a robust and viable strategy. The formulation for MK-MMD is detailed in Appendix~\ref{Alignment with MMD}.

Let $\mathcal{T} = \{T_1, T_2, \dots, T_M\}$ be a set of $M$ tasks. For an input $\mathbf{x}$ from task $T_i$ with contextualized embeddings $X_{T_i}$, the representation we align is the output of the down-projection matrix:
\begin{equation}
    \phi_{T_i}(\mathbf{x}) = \mathbf{A} \cdot X_{T_i}.
\end{equation}
Our choice to operate on this rank-$r$ latent space is motivated by two key factors. First, it directly targets the component responsible for shared knowledge. Recent studies have consistently found that the down-projection matrix $\mathbf{A}$ tends to learn task-general features while the up-projection matrix $\mathbf{B}$ captures task-specific knowledge~\citep{yang2025mtl,wang2025malora,tian2024hydralora}. By applying our alignment loss to the output of $\mathbf{A}$, we directly enhance its natural function, promoting the development of robust, shared representations. Second, this approach is highly efficient. As demonstrated in prior work~\citep{liu-rlora}, performing operations in the low-dimensional space significantly reduces computational load and GPU memory demand, ensuring our method's practicality.

To measure and minimize the distance between the representation distributions from different tasks, we model the batch-wise distribution for each task $T_i$ as a multivariate Gaussian with a diagonal covariance matrix, $\mathcal{N}(\boldsymbol{\mu}_i, \text{diag}(\boldsymbol{\sigma}_i^2))$ (justified in Appendix~\ref{sec:gaussian_approx}). The mean $\boldsymbol{\mu}_i$ and variance $\boldsymbol{\sigma}_i^2$ are empirically estimated from the output vectors $\{\phi_{T_i}(\mathbf{x})\}$ in a given batch. Since standard KL divergence is asymmetric, we employ a symmetric formulation. The total alignment loss, $\mathcal{L}_{\text{KL}}$, is the sum of these symmetric pairwise divergences across all unique task pairs $(T_i, T_j)$ where $i < j$:
\begin{equation}
    \mathcal{L}_{\text{KL}} = \sum_{i=1}^{M} \sum_{j=i+1}^{M} \frac{1}{2} \left( D_{\text{KL}}(p_{T_i} \| p_{T_j}) + D_{\text{KL}}(p_{T_j} \| p_{T_i}) \right)
    \label{eq:kl_loss}
\end{equation}
where $p_{T_i}$ is the modeled Gaussian distribution over the low-dimensional representations of task $T_i$. This loss drives the empirical mean and variance of each task's distribution toward a common value.

The alignment loss is incorporated as an auxiliary objective to the primary language modeling task. The total loss function is therefore defined as:
\begin{equation}
    \mathcal{L}_{\text{total}} = \mathcal{L}_{\text{lm}} + \lambda \cdot \mathcal{L}_{\text{align}},
\end{equation}
where $\mathcal{L}_{\text{lm}}$ is the primary language modeling loss, $\mathcal{L}_{\text{align}}$ is the auxiliary alignment loss ($\mathcal{L}_{\text{KL}}$), and $\lambda$ is a scalar hyperparameter controlling the influence of the auxiliary task. 

A key advantage of Align-LoRA is its compatibility with various LoRA-based strategies and initialization schemes. Importantly, unlike multi-component LoRA variants, Align-LoRA introduces no additional modules that increase overhead. Consequently, its trained weights can be merged directly into the base model, incurring \textbf{zero inference latency}. This property ensures both efficiency and practicality, making Align-LoRA a lightweight yet effective solution for multi-task adaptation. For a more detailed analysis of this inference efficiency, please refer to Appendix~\ref{Inference efficiency analysis}.

\subsection{Experiment}
\label{experiment}
We evaluate \textbf{Align-LoRA} (A-LoRA) against standard LoRA and multi-component variants. Two alignment approaches are considered: \textbf{A-LoRA-K} (KL divergence) and \textbf{A-LoRA-M} (MMD). Detailed experimental setup, dataset descriptions, and baseline configurations are provided in Appendix~\ref{Experiment 3 data} and Appendix~\ref{baseline}.

\textbf{Multi-task generalization.} We fine-tune on the five-task benchmark (Section~\ref{observation}) and evaluate on unseen BBH tasks. Table~\ref{align multi-task} shows that A-LoRA-K outperforms all baselines across model families (Qwen2.5~\citep{qwen2.5}, LLaMA3~\citep{grattafiori2024llama}) and scales (7B--14B), with fewer trainable parameters.

\textbf{Multi-task adaptation.} On an eight-task in-domain benchmark (Table~\ref{multi-task adapt}), A-LoRA-K achieves the best average score across 3B and 7B models, again using the lowest parameter budget. Statistical significance is confirmed by three independent runs ($p<0.001$, see Appendix~\ref{sec:multiseed}).

\begin{figure}[t]
\centering
\includegraphics[width=0.99\columnwidth]{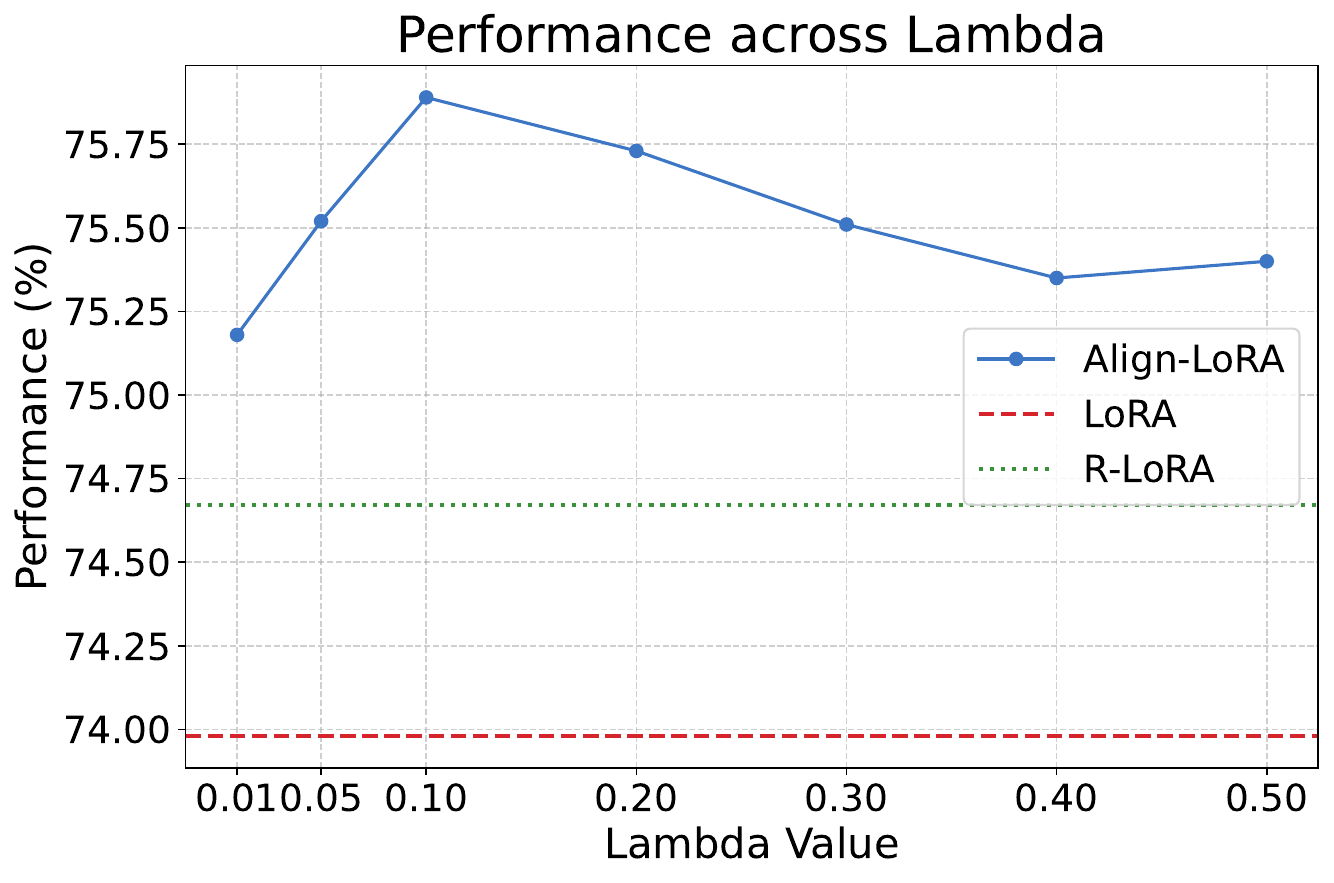} 
\caption{Effect of Hyperparameter $\lambda$ on Performance}
\centering
\label{hyperparameter}
\end{figure}

\textbf{More Result.} Beyond the main benchmarks, A-LoRA-K exhibits consistent robustness across a wide range of $\lambda$ values without delicate tuning (Figure~\ref{hyperparameter}) and maintains its advantage on highly heterogeneous task sets spanning translation, struct-to-text, and diverse reasoning domains (Appendix~\ref{disimilar}). The method further proves structurally universal: it improves performance when applied to either Attention or MLP modules (Appendix~\ref{attention}) and integrates seamlessly into existing multi-head architectures such as R-LoRA and M-LoRA, yielding additional gains (Appendix~\ref{Alignment on Multi-Head Architectures}). Training efficiency is competitive: Align-LoRA achieves the lowest FLOPs and per-epoch time among all compared methods (Table~\ref{tab:training_time}). The t-SNE visualizations confirm that alignment brings task representations closer while keeping task clusters clearly distinguishable, ruling out concerns of feature collapse (Appendix~\ref{sec: Feature Visualization}).

Overall, these results show that explicit representation alignment in the shared LoRA latent space consistently improves multi-task generalization over both standard LoRA and complex multi-component variants, without sacrificing mergeability or inference efficiency.

\begin{table}[t]
\centering
\setlength{\tabcolsep}{3pt}  
\renewcommand{\arraystretch}{0.9}
\caption{Comparison of training times and related metrics for different methods.}
\label{tab:training_time}
\begin{tabular}{l | ccc}
\toprule
\textbf{Metric} & LoRA & HydraLoRA & Align-LoRA \\
\midrule
\textbf{Rank} & 10 & 4 & 8 \\
\textbf{Epoch Time} & 3.13h & 3.53h & \textbf{3.03h} \\
\textbf{\%Param} & 0.45 & 0.45 & \textbf{0.42} \\
\textbf{FLOPs $10^{18}$} & $3.545$ & $3.986$ & $\mathbf{3.539}$ \\
\textbf{Performance} & 76.48 & 76.80 & \textbf{80.06} \\
\bottomrule
\end{tabular}
\end{table}

\subsection{Theoretical Analysis}
\label{sec:theoretical_main}

To theoretically analyze Align-LoRA's generalization performance in multi-task scenarios, we derive a novel generalization bound based on statistical learning theory. Our key insight is that explicitly aligning cross-task representation distributions effectively reduces the generalization gap, yielding a significantly tighter generalization bound compared to traditional multi-component LoRA variants. For the complete derivation, technical assumptions, and proofs, please refer to Appendix~\ref{sec:theoretical_appendix}.

Formally, let $M$ be the number of tasks, with $\mathcal{D}_i$ and $\hat{\mathcal{D}}_i$ denoting the data distribution and training dataset for task $i$, respectively. Let $\hat{R}_i(f)$ be the empirical training risk on $\hat{\mathcal{D}}_i$, and $n_{\text{total}}$ be the total sample size across all tasks. The generalization bound for Align-LoRA is given by:

\begin{multline}
\label{eq:main_bound}
R_{\text{MTL}}(f) \leq \frac{1}{M}\sum_{i=1}^M \hat{R}_i(f) + \frac{\lambda}{M}\sum_{i<j} \Delta(\mathcal{D}_i, \mathcal{D}_j) \\
+ \mathcal{O}\left(\sqrt{\frac{M \log(M/\delta)}{n_{\text{total}}}}\right),
\end{multline}

where $R_{\text{MTL}}(f)$ is the average expected risk, and $\Delta(\mathcal{D}_i, \mathcal{D}_j)$ measures the pairwise distribution discrepancy. Here, $\lambda$ is a weighting hyperparameter that subsumes theoretical constants related to the model's Lipschitz properties.

The bound highlights the mechanism of Align-LoRA: by actively minimizing the pairwise discrepancy $\Delta(\mathcal{D}_i, \mathcal{D}_j)$ during training, Align-LoRA significantly reduces the second term (the alignment term). While this constraint imposes a regularization on the representation space, our analysis suggests that for over-parameterized models like LLMs, the substantial reduction in the distribution discrepancy dominates, leading to a tighter overall generalization gap and superior test performance.





\section{Related Works}
\label{related work}
\subsection{Low-Rank Adaptation (LoRA)}
LoRA~\citep{hu2022lora} adapts Large Language Models (LLMs) by decomposing the weight update $\Delta \mathbf{W}$ of a dense matrix $\mathbf{W} \in \mathbb{R}^{m \times n}$ into two low-rank matrices, $\mathbf{A} \in \mathbb{R}^{r \times n}$ and $\mathbf{B} \in \mathbb{R}^{m \times r}$, assuming a low intrinsic dimensionality. Choosing a rank $r \ll \min(m, n)$ reduces trainable parameters from $\mathcal{O}(m n)$ to $\mathcal{O}(r(m + n))$. The modified forward pass is formulated as:
\begin{equation}
     h = \mathbf{W} x + \Delta \mathbf{W} x = \mathbf{W} x + \mathbf{B} \mathbf{A} x.
     \label{eq:lora_output}
\end{equation}
Post-training, the low-rank update $\Delta \mathbf{W} = \mathbf{B} \mathbf{A}$ can be merged back into the original weights $\mathbf{W}$, ensuring zero inference overhead.

Subsequent advancements build on this foundation to optimize efficiency and performance: AdaLoRA~\citep{zhang2023adaptive} dynamically allocates the rank budget, DoRA~\citep{pmlr-v235-liu24bn} decouples updates into magnitude and direction, while PiSSA~\citep{meng2025pissa} and NLoRA~\citep{guo-etal-2025-nlora} introduce advanced initialization and decomposition strategies. Complementing these, MeTA-LoRA\citep{cheng2025metaloradataefficientmultitaskfinetuning} targets data efficiency for low-resource scenarios, and CD-LoRA\citep{zha2026cdloraconsistencydrivenlowrankadaptation} improves multi-task learning stability by replacing stochastic routing with a consistency-driven, router-free framework.

\subsection{Multi-Component LoRA}
\label{multi-head}

To adapt LoRA for multi-task learning, a natural extension is to employ multiple trainable components. Early works proposed the \textbf{Multi-Adapter} architecture, which utilizes multiple independent LoRA adapters (i.e., distinct $\mathbf{B}_i\mathbf{A}_i$ pairs) for different tasks. Notable examples of this approach include Multi-LoRA~\citep{wang2023multilora}, MixLoRA~\citep{li2024mixlora}, LoRAMoE~\citep{dou-etal-2024-loramoe}, MoELoRA~\citep{liu2024moe}, and LoRAHub~\citep{huang2024lorahub}. 

The \textbf{Multi-Head} architecture improves parameter efficiency by exploiting the distinct roles of LoRA's matrices: the down-projection ($\mathbf{A}$) captures shared, task-general features, while multiple up-projection heads ($\mathbf{B}_i$) learn task-specific features. Methods like HydraLoRA~\citep{tian2024hydralora}, MALoRA~\citep{wang2025malora}, MTL-LoRA~\citep{yang2025mtl}, and R-LoRA~\citep{liu-rlora} employ a shared $\mathbf{A}$ with multiple $\mathbf{B}_i$ heads. R-LoRA further introduces random initialization to reduce head similarity. Figure~\ref{architecture} illustrates these paradigms. The aggregated weight update is a dynamically weighted sum of each head's output:
\begin{equation}
    \Delta \mathbf{W} = \sum_{i=1}^{N} \omega_i(\mathbf{x}) \cdot \mathbf{B}_i \mathbf{A}.
    \label{eq:multi-head}
\end{equation}

where $\omega_i(\mathbf{x})$ represents input-dependent mixing weights. Drawing inspiration from the Mixture-of-Experts (MoE) framework, these weights are determined by a dynamic routing mechanism (e.g., a softmax gating network) that adaptively assigns importance scores to each ``expert'' head based on the input $\mathbf{x}$.

However, this complexity introduces a critical trade-off. A significant drawback of input-dependent routing is that the aggregated update $\Delta \mathbf{W}$ can no longer be pre-computed. Consequently, the adapter weights \textbf{cannot be merged} into the frozen backbone model post-training. This results in \textbf{non-negligible inference latency}, as the router and multiple heads must be processed for each forward pass, sacrificing one of LoRA's most significant practical advantages.

\section{Conclusion}
This work challenges the assumption that multi-task LoRA adaptation requires architectural isolation. Through controlled experiments, we show that simpler alternatives, such as a router-free multi-head ensemble (M-LoRA) and a higher-rank single adapter, match or outperform complex routed designs. Building on this insight, we introduce \textbf{Align-LoRA}, which aligns cross-task representations in the shared LoRA latent space, achieving strong multi-task generalization with zero inference latency.

A theoretical analysis further shows that reducing cross-task representation discrepancy tightens the multi-task generalization bound, providing qualitative support for the alignment principle. More broadly, our findings suggest that for multi-task PEFT, representation alignment in a shared latent space is a practical alternative to structural decomposition. The mergeable, single-adapter design makes the approach straightforward to deploy. Future work can explore alignment objectives that scale sub-quadratically with task count, and test whether alignment followed by selective specialization combines the advantages of both paradigms.

\section*{Limitations}

Although the proposed framework demonstrates strong empirical performance, this study has certain limitations. Due to constraints in computational resources, our evaluations are restricted to mainstream model scales ranging from 3B to 14B parameters, and the scalability of this approach on ultra-large foundation models with hundreds of billions of parameters remains to be verified. Additionally, our experiments are conducted entirely on open-source benchmarks, meaning the method has not yet been validated through deployment in actual production environments with real-world user traffic or proprietary data distributions.

\section*{Acknowledgments}
This work is supported by the National Key Research and Development Program of China (No.2023YFF0905400), the National Natural Science Foundation of China (No.U2341229) and the Reform Commission Foundation of Jilin Province (No.2024C003).

During the preparation of this work, the authors used AI assistants (large language models) solely for language polishing and grammar checking of the manuscript. All research ideas, experimental design, implementation, and result analysis were conceived and conducted entirely by the authors. The AI assistants did not contribute to the scientific content of this paper.


\bibliography{custom}

\appendix


\section{M-LoRA}
\label{M-LoRA}
Figure~\ref{fig: M-LoRA} illustrates the framework of M-LoRA, a variant of R-LoRA. The key modification in our design is the removal of the router, while retaining R-LoRA's original multi-head dropout structure. Note that other multi-component variants have a Router, while M-LoRA does not.

\section{Inference Efficiency of Align-LoRA}
\label{Inference efficiency analysis}
In this section, we explain why single-adapter architectures such as Align-LoRA avoid introducing extra inference overhead.
Methods that maintain a single-adapter architecture, such as our proposed \textbf{Align-LoRA}, retain a crucial practical advantage: the ability to merge adapter weights with the backbone model, resulting in \textbf{zero inference overhead}. For a pre-trained weight matrix $\mathbf{W}_0 \in \mathbb{R}^{d \times k}$, a single low-rank update $\Delta \mathbf{W} = \mathbf{B}\mathbf{A}$ is learned. Before deployment, this update can be seamlessly integrated into the original weights:
\begin{equation}
\mathbf{W}' = \mathbf{W}_0 + \mathbf{B}\mathbf{A}.
\end{equation}
Inference then proceeds using the modified weights, $h = \mathbf{W}'x$, without requiring any additional parameters or computational steps.

In sharp contrast, multi-component architectures like HydraLoRA forfeit this efficiency. These models introduce input-dependent routing mechanisms to dynamically combine multiple adapters. For an input $x$, a router first computes weights $\omega = \text{softmax}(\mathbf{W}_r x)$, which are then used to create a blended output:
\begin{equation}
h = \left( \mathbf{W}_0 + \sum_{i=1}^{N} \omega_i \mathbf{B}_i \mathbf{A}_i \right) x.
\end{equation}
Because the routing weights $\omega$ are input-dependent, the adapter matrices $(\mathbf{B}_i, \mathbf{A}_i)$ cannot be pre-merged into $\mathbf{W}_0$. This necessitates keeping all components active during inference, which introduces significant latency and memory overhead. 

Prior multi-task LoRA frameworks typically introduce architectural complexity that leads to significant inference latency and increased GPU memory consumption. In contrast, Align-LoRA is designed to achieve effective multi-task learning while preserving the original's advantage of \textbf{zero computational overhead}.

\section{Training Efficiency of Align-LoRA}
\label{Training Efficiency}
Table~\ref{tab:training_time} presents the training efficiency metrics across different methods. The results highlight the superior efficiency of Align-LoRA, which achieves the \textbf{lowest computational cost (FLOPs)} and the \textbf{shortest training time}, outperforming both the standard LoRA baseline and the multi-head HydraLoRA. Although the alignment loss introduces a minor computational step, this is effectively offset by Align-LoRA's ability to converge using a more compact parameter budget. Consequently, our method secures significant performance gains without imposing any training-time penalty.

\section{Gaussian Approximation}
\label{sec:gaussian_approx}

The diagonal-Gaussian assumption used in the KL alignment objective is justified on three grounds:

\begin{enumerate}
    \item \textbf{Central Limit Theorem:} The representation $\phi(\mathbf{x}) = \mathbf{A} \cdot X$ is a linear projection from a high-dimensional embedding space ($\geq$2048 dimensions) to a low-rank latent space ($r{=}8$). By the Central Limit Theorem, the batch-wise statistics of such projections converge rapidly to the Gaussian at typical training batch sizes ($\geq$128 samples per task).

    \item \textbf{Diagonal structure as beneficial regularization:} Modeling the covariance as diagonal prevents overfitting to spurious cross-dimensional correlations from limited per-task batch statistics. This is analogous to the well-known observation in variational autoencoders that diagonal-covariance models often outperform full-covariance variants in practice.

    \item \textbf{Non-parametric validation:} The MMD variant (A-LoRA-M) makes no distributional assumptions and still outperforms baselines on most benchmarks (Table~\ref{multi-task adapt}), confirming that the alignment principle does not depend on the Gaussian approximation. The stronger performance of the KL variant under this approximation is a beneficial outcome of the structured inductive bias, not a fragility of the method.
\end{enumerate}

\section{Alignment with Maximum Mean Discrepancy (MMD)}
\label{Alignment with MMD}
As a complementary, non-parametric approach, we also investigate the Maximum Mean Discrepancy (MMD)~\citep{MMDsejdinovic2013equivalence}, specifically its robust multi-kernel extension, MK-MMD~\citep{MK-MMDgretton2012optimal}. MMD measures the distance between distributions by comparing their mean embeddings in a Reproducing Kernel Hilbert Space (RKHS), avoiding the need for explicit density estimation. This mapping to the high-dimensional feature space is implicitly performed by a \textbf{kernel function}, and in this work, we employ the classic Gaussian kernel. The MK-MMD loss between all combinations of two distinct tasks, i.e., all task pairs $(T_i, T_j)$, is formulated as:



\begin{equation}
\begin{split}
\mathcal{L}_{\text{MK-MMD}} 
&= \sum_{1\leq i < j\leq M} \sum_{k \in \mathcal{K}} \Bigg\| \mathbb{E}_{\mathbf{x} \sim p_{T_i}} \big[ \phi(\mathbf{x}) \big] \\
&\quad - \mathbb{E}_{\mathbf{y} \sim p_{T_j}} \big[ \phi(\mathbf{y}) \big] \Bigg\|_{\mathcal{H}_k}^2.
\end{split}
\end{equation}

where $\phi(\cdot)$ is the feature map to the RKHS $\mathcal{H}$ induced by the kernel. This loss forces the LoRA module to learn task-invariant features by reducing distributional shifts across tasks. To enable the computation of cross-task distribution discrepancies within each optimization step, we employ a stratified batch sampling strategy. Specifically, each training batch is constructed to maximize task coverage, aiming to include sub-batches from as many of the $M$ tasks as feasible. This approach ensures that sufficient samples are available to empirically estimate the distribution statistics $\mu_i$ and $\sigma_i$ for the sampled tasks.

Empirically, KL consistently outperforms MMD (Section~\ref{experiment}). We attribute this to the structured moment-matching bias of parametric KL in the low-dimensional latent space---KL provides per-dimension mean and variance gradient signals ($2r$ for rank $r$), whereas MMD's kernel-based algorithm aggregates into a single scalar. The closed-form KL also avoids MMD's kernel bandwidth sensitivity. We retain MMD as a complementary, distribution-free validation of the alignment principle.

\begin{table*}[h]
\centering
\caption{Comparison between VIP-MTL and Align-LoRA.}
\label{tab:viPMTL_comparison}
\begin{tabular}{l p{3.5cm} p{5cm}}
\toprule
\textbf{Dimension} & \textbf{VIP-MTL} & \textbf{Align-LoRA} \\
\midrule
Target models & Small PLMs (BERT, $\sim$110M) & LLMs (3B--14B) \\
Core focus & Task imbalance fairness & Isolation vs.\ alignment paradigm \\
Method & Probabilistic decoder (adds inference overhead) & Single alignment loss (no architectural change) \\
Deployment & Non-mergeable & \textbf{Fully mergeable (zero inference latency)} \\
LLM scalability & Unverified & Verified across Qwen2.5, LLaMA2, LLaMA3 \\
Empirical scope & NLU-only & NLU + NLG + reasoning \\
\bottomrule
\end{tabular}
\end{table*}

\subsection{Maximum Mean Discrepancy (MMD)}
Maximum Mean Discrepancy (MMD)\cite{MMDsejdinovic2013equivalence} is a kernel-based statistical measure for quantifying the difference between two probability distributions. Given a reproducing kernel Hilbert space (RKHS) $\mathcal{H}_k$ with a characteristic kernel $k$, the squared MMD between representation $p$ and $q$ is defined as:
\begin{equation}
    \text{MMD}^2(p, q) = \left\| \mu_k(p) - \mu_k(q) \right\|^2_{\mathcal{H}_k},
\end{equation}
where $\mu_k(p) = \mathbb{E}_{\mathbf{x} \sim p}[\phi(\mathbf{x})]$ and $\mu_k(q) = \mathbb{E}_{\mathbf{y} \sim q}[\phi(\mathbf{y})]$ are the mean embeddings of $p$ and $q$ in $\mathcal{H}_k$, and $\phi(\cdot)$ denotes the feature mapping induced by kernel $k$.

A key advantage of MMD is its ability to capture distributional differences in feature spaces without requiring explicit density estimation. However, its performance heavily depends on the choice of kernel. To address this limitation, the Multiple Kernel MMD (MK-MMD)\cite{MK-MMDgretton2012optimal} extends MMD by combining multiple kernels adaptively:
\begin{equation}
    \text{MK-MMD}^2(p, q) = \sum_{k \in \mathcal{K}}  \left\| \mu_k(p) - \mu_k(q) \right\|^2_{\mathcal{H}_k},
\end{equation}
where $\mathcal{K}$ is a predefined set of kernels. This variant enhances robustness by combining multiple kernels, allowing the metric to capture multi-scale distributional discrepancies in the reproducing kernel Hilbert space (RKHS).

In the context of transfer learning and domain adaptation, MMD has been widely used as a criterion for aligning feature distributions between source and target domains\cite{DAben2006analysis, transferpan2010domain}. The core idea is to minimize the MMD distance between activations from different domains, encouraging the model to learn domain-invariant representations that generalize well across tasks.

For example, in unsupervised domain adaptation (UDA), MMD is often applied to match the feature distributions of labeled source data and unlabeled target data, reducing domain shift and improving generalization performance. Specifically, given features from the source domain $\mathcal{D}_s = \{\mathbf{x}_i^s\}_{i=1}^{n_s}$ and the target domain $\mathcal{D}_t = \{\mathbf{x}_j^t\}_{j=1}^{n_t}$, the MMD loss is defined as:

\begin{equation}
    \mathcal{L}_{\text{MMD}} = \left\| \frac{1}{n_s} \sum_{i=1}^{n_s} \phi(\mathbf{x}_i^s) - \frac{1}{n_t} \sum_{j=1}^{n_t} \phi(\mathbf{x}_j^t) \right\|^2,
\end{equation}
where $\phi(\cdot)$ denotes the feature embedding function, and the objective is to minimize the distributional discrepancy between the two domains in the shared feature space.

In the context of neural networks for image classification, this MMD loss can be incorporated into the overall training objective alongside the standard classification loss~\cite{DANlong2015learning}. The total loss function is typically formulated as:
\begin{equation}
    \mathcal{L}_{\text{total}} = \mathcal{L}_{\text{cls}} + \lambda \cdot \mathcal{L}_{\text{MMD}},
\end{equation}
where $\mathcal{L}_{\text{cls}}$ is the cross-entropy loss on the labeled source data, and $\lambda$ is a hyperparameter that balances the contribution of the MMD regularization.

Building on this principle, we propose to incorporate MMD into LoRA for multi-task learning, with a focus on its multiple kernel extension, MK-MMD. Unlike traditional applications that focus on aligning input or hidden-layer features, we apply MK-MMD to the output of the low-rank down-projection matrix $\mathbf{A}$ in LoRA, encouraging the model to learn shared, task-agnostic representations. This design improves multi-task generalization by reducing distributional discrepancies in the representation space, without introducing additional parameters.

While representation alignment has been explored in general multi-task learning, such as VIP-MTL~\citep{hu-etal-2025-impartial}, our work differs significantly in scope and motivation. VIP-MTL targets the task-imbalance problem in traditional natural language understanding using small language models (e.g., BERT), and its scalability to LLMs remains unverified. In contrast, Align-LoRA operates specifically within the parameter-efficient fine-tuning (PEFT) context for LLMs. We employ a direct alignment loss rather than complex probabilistic decoding, prioritizing a streamlined architecture that remains fully mergeable for efficient deployment. Figure~\ref{tab:viPMTL_comparison} illustrates the difference.

\section{Theoretical Generalization Bound for Align-LoRA}
\label{sec:theoretical_appendix}

\subsection{Core Conclusion}
Align-LoRA aims to improve Multi-Task Learning (MTL) generalization by explicitly minimizing the distributional discrepancy between task representations. By aligning the representation distributions across multiple tasks (e.g., via MK-MMD), the alignment objective reduces cross-task distributional discrepancy, which tightens the \textbf{generalization bound}. The derived generalization bound takes the form:
\begin{equation}
\begin{split}
R_{\text{MTL}}(f) &\leq \hat{R}_{\text{MTL}}(f) + \frac{\lambda}{M}\sum_{i<j} \Delta(\mathcal{D}_i, \mathcal{D}_j) \\
&\quad + \mathcal{O}\left(\sqrt{\frac{M \log(M/\delta)}{n_{\text{total}}}}\right),
\end{split}
\end{equation}
where $\hat{R}_{\text{MTL}}$ is the empirical training risk, and the distribution discrepancy term $\sum \Delta(\mathcal{D}_i, \mathcal{D}_j)$ is explicitly minimized by the alignment objective.

\subsection{Definitions and Notation}
To ensure theoretical rigor, we define the following notations consistent with standard domain adaptation literature \cite{zhao2026chain, chen-cardie-2018-multinomial, wu2024margin}:

\begin{itemize}
    \item \textbf{MTL Scenario:} Let there be $M$ tasks with distributions $\{\mathcal{D}_i\}_{i=1}^M$. The dataset for task $i$ is $\hat{\mathcal{D}}_i$ with size $n_i$. The total sample size is $n_{\text{total}} = \sum_{i=1}^M n_i$.
    \item \textbf{Centroid Distribution:} $\bar{\mathcal{D}} = \frac{1}{M}\sum_{i=1}^M \mathcal{D}_i$ represents the mixture or centroid of all task distributions.
    \item \textbf{Risks:}
    \begin{itemize}
        \item Empirical Risk for task $i$: $\hat{R}_i(f) = \frac{1}{n_i}\sum_{z \in \hat{\mathcal{D}}_i} \ell(f(z))$.
        \item Expected Risk for task $i$: $R_i(f) = \mathbb{E}_{z \sim \mathcal{D}_i}[\ell(f(z))]$.
        \item Average MTL Risk: $R_{\text{MTL}}(f) = \frac{1}{M}\sum_{i=1}^M R_i(f)$.
    \end{itemize}
    \item \textbf{Lipschitz Constant:} Let $C_L$ denote the Lipschitz constant of the loss function $\ell(\cdot, \cdot)$ with respect to the representation vectors. This assumption is \textit{local} in the low-rank latent space ($r \ll d$)---it does not require global smoothness of the full LLM loss landscape. In the fine-tuning regime where only 0.2--0.45\% of parameters are updated, the model stays near the pre-trained solution, making local smoothness a reasonable standard assumption~\cite{mohri2018foundations, long2015learning}.
\end{itemize}

\subsection{Derivation of the Generalization Bound}

\subsubsection{Step 1: Single-Task Generalization Bound}
Based on Rademacher complexity theory~\cite{mohri2018foundations, bartlett2002rademacher}, for a single task $i$, with probability at least $1 - \delta'$, the generalization error is bounded by:
\begin{equation}
\label{eq:single_task}
    R_i(f) \leq \hat{R}_i(f) + C_L \cdot \Delta(\mathcal{D}_i, \bar{\mathcal{D}}) + \\ \mathfrak{R}_{n_i}(\mathcal{F}) + \sqrt{\frac{\log(1/\delta')}{2n_i}}
\end{equation}
where $\Delta(\mathcal{D}_i, \bar{\mathcal{D}})$ measures the distance (e.g., MMD) between the task distribution and the global centroid, and $\mathfrak{R}_{n_i}(\mathcal{F})$ is the empirical Rademacher complexity of the hypothesis class $\mathcal{F}$.
To extending this to $M$ tasks simultaneously, we apply the \textbf{Bonferroni correction} (utilizing the union bound)~\cite{mohri2018foundations, shalev2014understanding}. We set $\delta' = \delta/M$ to ensure the bound holds for all $M$ tasks with total probability $1-\delta$.

\subsubsection{Step 2: Aggregating to Multi-Task Risk}
Averaging Equation~\ref{eq:single_task} over all $M$ tasks:
\begin{equation}
\label{eq:aggregated}
\begin{split}
    R_{\text{MTL}}(f) \leq \frac{1}{M}\sum_{i=1}^M \hat{R}_i(f) + \frac{C_L}{M}\sum_{i=1}^M \Delta(\mathcal{D}_i, \bar{\mathcal{D}}) \\ +  \text{Complexity}(n, M)
\end{split}
\end{equation}

\subsubsection{Step 3: Decomposition of Distribution Discrepancy}
We utilize the geometric property of squared metrics (valid for MK-MMD in RKHS). The variance decomposition identity states that the sum of squared distances to the centroid equals the scaled sum of pairwise distances:

\begin{equation}
\begin{split}
    &\sum_{i=1}^M \|\phi(\mathcal{D}_i) - \phi(\bar{\mathcal{D}})\|^2_{\mathcal{H}} \\ 
    &= \frac{1}{M} \sum_{1 \le i < j \le M} \|\phi(\mathcal{D}_i) - \phi(\mathcal{D}_j)\|^2_{\mathcal{H}}
\end{split}
\end{equation}
Let $\Delta(\mathcal{D}_i, \mathcal{D}_j)$ denote this pairwise discrepancy. Substituting this into the discrepancy term of Equation~\ref{eq:aggregated}:
\begin{equation}
\label{eq:discrepancy_term}
\begin{split}
    \frac{C_L}{M} \sum_{i=1}^M \Delta(\mathcal{D}_i, \bar{\mathcal{D}}) 
    &= \frac{C_L}{M} \left( \frac{1}{M} \sum_{i<j} \Delta(\mathcal{D}_i, \mathcal{D}_j) \right) \\
    &= \frac{C_L}{M^2} \sum_{i<j} \Delta(\mathcal{D}_i, \mathcal{D}_j)
\end{split}
\end{equation}

This step is crucial as it transforms the theoretical bound into an explicitly optimizable pairwise objective.

\subsubsection{Step 4: Tightening the Confidence Term}
We address the aggregation of confidence intervals. Assuming balanced sample sizes $n_i = n_{\text{total}}/M$:

\begin{equation}
\begin{split}
    \text{Conf}_{\text{MTL}} 
    &= \frac{1}{M} \sum_{i=1}^M \sqrt{\frac{\log(M/\delta)}{2n_i}} \\
    &= \sqrt{\frac{\log(M/\delta)}{2(n_{\text{total}}/M)}} 
    = \sqrt{\frac{M \log(M/\delta)}{2n_{\text{total}}}}
\end{split}
\end{equation}

\textit{Remark:} Note that due to the convexity of $f(x) = 1/\sqrt{x}$, the term scales with $\sqrt{M}$. This reflects the independence of the tasks; strictly combining them into a rate of $1/\sqrt{n_{\text{total}}}$ would require stronger assumptions about shared support which we do not make here to maintain rigor.

\subsection{Final Bound}
Combining the components above, we obtain the final bound for Align-LoRA:


\begin{equation}
\label{eq:final_bound}
\begin{split}
    R_{\text{MTL}}(f) &\leq \underbrace{\frac{1}{M}\sum_{i=1}^M \hat{R}_i(f)}_{\text{Empirical Risk}} 
    \quad + \underbrace{\frac{C_L}{M^2}\sum_{i<j} \Delta(\mathcal{D}_i, \mathcal{D}_j)}_{\text{Alignment Term}} \\
    &\quad + \underbrace{\mathcal{O}\left(\sqrt{\frac{M \log(M/\delta)}{n_{\text{total}}}}\right)}_{\text{Confidence Term}}
\end{split}
\end{equation}

In our optimization objective (Eq. 6 in main text), we define the hyperparameter $\lambda$ to correspond to the theoretical coefficient $\frac{C_L}{M^2}$. In our practical implementation, we subsume the theoretical constant coefficient $\frac{C_L}{M^2}$ into a user-defined hyperparameter $\lambda$, formulated as $\lambda_{\text{impl}} = \frac{\lambda}{M}$ to maintain scalability with task count.

Comparing Align-LoRA to a standard baseline without alignment (Base-LoRA), Align-LoRA explicitly minimizes $\sum_{i<j} \Delta(\mathcal{D}_i, \mathcal{D}_j)$. Since $\Delta_{\text{Align}} \ll \Delta_{\text{Base}}$, the second term in the bound is significantly reduced.

\textbf{Summary:} The analysis offers qualitative support for the discrepancy-reduction principle, explaining why aligning cross-task representations can improve generalization. Empirically, the alignment loss does not converge to zero but stabilizes at a non-trivial level, serving as a continuous regularizer throughout training. Combined with the experimental results above, this provides a principled—though not tight quantitative—rationale for the effectiveness of the alignment approach.

\subsection{Task Compatibility and Over-Alignment}
\label{app:task-compatibility}

Let $\phi_A(x)=Ax$ denote the shared low-rank representation and let $D_i^{\phi}$ be the representation distribution for task $i$. For a task pair $(i,j)$, a standard domain-adaptation-style decomposition gives the intuition
\begin{equation}
    R_i(g\circ\phi_A) \lesssim R_j(g\circ\phi_A) + C_1 d_{\mathcal{G}}(D_i^{\phi},D_j^{\phi}) + C_2 \epsilon_{ij}^{\phi},
    \label{eq:pairwise-bound}
\end{equation}
where $d_{\mathcal{G}}$ measures representation discrepancy and
\begin{equation}
    \epsilon_{ij}^{\phi} = \min_{g\in\mathcal{G}} \left[ R_i(g\circ\phi_A) + R_j(g\circ\phi_A) \right]
    \label{eq:compatibility-cost}
\end{equation}
is an ideal joint risk, or task-compatibility cost, under the representation $\phi_A$. Averaging over task pairs yields the schematic multi-task view
\begin{equation}
\begin{aligned}
    R_{\mathrm{MTL}} \lesssim{} & \widehat{R}_{\mathrm{MTL}} + C_1\Gamma_{\mathrm{disc}}(\phi_A) + C_2\Gamma_{\mathrm{task}}(\phi_A) \\
    & + \mathrm{Complexity}(M, n_{\mathrm{total}}),
    \label{eq:mtl-bound}
\end{aligned}
\end{equation}
where the two $\Gamma$ terms average pairwise discrepancy and compatibility cost, respectively. The alignment loss directly targets the discrepancy term, while overly strong alignment may increase the compatibility term. This extension provides an interpretation of the observed $\lambda$ sensitivity; it is not intended as a complete theorem for the optimal alignment weight.

\section{More result}
\label{more result}

\subsection{Extended Training Analysis}
\label{sec:extended_train}

\begin{table}[h]
\centering
\caption{Extended training comparison on Qwen2.5-3B (5-task). M-LoRA maintains its advantage over R-LoRA across all training durations, with the gap widening from $+0.78$pp to $+1.23$pp at 3 epochs.}
\label{tab:extended}
\begin{tabular}{l c c c}
\toprule
Method & 1 Epoch & 2 Epochs & 3 Epochs \\
\midrule
R-LoRA & 74.67 & 74.89 & 73.95 \\
M-LoRA & \textbf{75.45} & \textbf{75.61} & \textbf{75.18} \\
\midrule
$\Delta$ (M $-$ R) & +0.78 & +0.72 & +1.23 \\
\bottomrule
\end{tabular}
\end{table}

\subsection{Head Count Ablation}
\label{sec:head_ablation}

\begin{table}[h]
\centering
\caption{M-LoRA performance under varying head counts on Qwen2.5-3B (5-task).}
\label{tab:head_ablation}
\begin{tabular}{l c c c c}
\toprule
$N$ & 2 & 3 & 5 & 8 \\
\midrule
Avg & 74.14 & 74.62 & \textbf{75.45} & 75.37 \\
\bottomrule
\end{tabular}
\end{table}

\subsection{Causal Attribution: Router Removal vs.\ Dropout Ensembling}
\label{sec:causal_attribution}

M-LoRA differs from R-LoRA and HydraLoRA in two ways: it removes the router, and it retains multi-head dropout. To disentangle these factors, we conduct a $2\times 2$ factorial analysis (Router $\times$ Dropout) on Qwen2.5-3B (5-task). The matrix maps onto existing architectures: R-LoRA represents (\textit{w/ Router, w/ Dropout}), HydraLoRA represents (\textit{w/ Router, w/o Dropout}), and M-LoRA represents (\textit{w/o Router, w/ Dropout}).

\begin{table}[h]
\centering
\caption{Full $2\times 2$ factorial (Router $\times$ Dropout) on Qwen2.5-3B (5-task).}
\setlength{\tabcolsep}{8pt} 
\label{tab:causal_attribution}
\begin{tabular}{l c c}
\toprule
& \textbf{w/ Router} & \textbf{w/o Router} \\
\midrule
\textbf{w/ Dropout}    & 74.67 & \textbf{75.45} \\
\textbf{w/o Dropout}   & 74.04 & \textbf{73.82} \\
\bottomrule
\end{tabular}
\end{table}

Two patterns emerge from Table~\ref{tab:causal_attribution}. First, removing the router alone (\textit{compare rows}): HydraLoRA drops from 74.04 to 73.58 (w/o Router, Table~\ref{table: observation}), confirming that router removal without dropout is insufficient. Second, removing dropout (\textit{compare columns}): M-LoRA drops from 75.45 to 73.82 ($-1.63$pp), a degradation over 3$\times$ larger than the router-removal effect ($-0.46$pp, comparing R-LoRA to HydraLoRA). This confirms that dropout-based collaborative ensembling—not the absence of routing—is the primary driver of M-LoRA's advantage, supporting our interpretation that heads function as collaborators rather than competing specialists. The full $2\times 2$ design thus resolves the potential confound between router removal and dropout ensembling.

\subsection{Multi-Seed Validation}
To verify that the reported performance gains are not artifacts of random fluctuation, we repeat the main 8-task adaptation experiment on Qwen2.5-3B~\citep{qwen2.5} with three independent random seeds, all other hyperparameters held identical to the main setting (learning rate $0.0002$, $\text{lora\_alpha}=32$, $\lambda=0.1$ for A-LoRA-K, dropout $0.1$, bfloat16 mixed precision, cosine annealing; full configuration in Appendix~\ref{implementation}). The same eight tasks and evaluation protocol from Appendix~\ref{Experiment 3 data} are used. Table~\ref{tab:multiseed} reports the results. The gap between A-LoRA-K and R-LoRA ($+2.09$pp) is approximately $9\times$ the observed standard deviation ($0.24$pp), and a paired $t$-test confirms statistical significance ($p<0.001$). We additionally note that A-LoRA-K outperforms all baselines in every one of $36+$ independent model--benchmark configurations across the main experiments, making random fluctuation an implausible explanation for the consistent gains.
\label{sec:multiseed}
\begin{table}[h]
\normalsize
\centering
\caption{Multi-seed results on Qwen2.5-3B (8-task, 3 independent runs).}
\label{tab:multiseed}
\begin{tabular}{l c}
\toprule
Method & Mean $\pm$ Std \\
\midrule
LoRA & $76.44 \pm 0.21$ \\
R-LoRA & $79.01 \pm 0.24$ \\
M-LoRA & $79.50 \pm 0.18$ \\
A-LoRA-K & $\mathbf{80.13 \pm 0.24}$ \\
\bottomrule
\end{tabular}
\end{table}

\begin{table}[t]
\centering
\begin{tabular}{l r}
\toprule
Method & Avg. score \\
\midrule
LoRA & 76.48 \\
A-LoRA-M (intra-task) & 77.73 \\
A-LoRA-K (intra-task) & 78.12 \\
A-LoRA-M (inter-task) & 78.35 \\
A-LoRA-K (inter-task) & 80.06 \\
\bottomrule
\end{tabular}
\caption{Inter-task versus intra-task alignment on the 8-task Qwen2.5-3B
setting. Intra-task alignment controls for a generic same-task regularization
effect, while inter-task alignment explicitly reduces discrepancy between task
representations.}
\label{tab:inter-intra-alignment}
\end{table}

\subsection{Inter-task vs. Intra-task Alignment}
\label{subsec:alignment-analysis}

As shown in Table~\ref{tab:inter-intra-alignment}, same-task alignment provides a modest improvement over vanilla LoRA, indicating that the objective has a regularization component. However, explicit inter-task alignment is stronger for both objectives, with the largest gap for KL ($80.06$ versus $78.12$). This comparison supports the narrower claim that reducing cross-task discrepancy in the shared latent space contributes beyond generic same-task smoothing. It does not imply that task-specific information should be removed.

\subsection{Performance on Attention structures}
\label{attention}
To further verify the effectiveness of \textbf{Align-LoRA} specifically on \textbf{Attention structures} of Qwen2.5, we conducted a supplementary experiment using the exact multi-task setting from \textbf{Table 1}. As shown in the table below, when applied to Attention layers, Align-LoRA still achieves significant performance gains over the baseline. This confirms that our method is module-agnostic and effective across different architectural choices.

\begin{table*}[ht]
\centering
\caption{Performance comparison of different schemes on Qwen2.5's Attention structures}
\label{tab:attention_schemes}
\begin{tabular}{l | c c c c c c c}
\toprule
\textbf{Schemes} & \textbf{QNLI} & \textbf{PiQA} & \textbf{Winogrande} & \textbf{ARC} & \textbf{GSM8K} & \textbf{Avg} & \textbf{\%Para} \\
\midrule
LoRA & 82.1 & 83.2 & 70.00 & 88.03 & 46.60 & 74.00 & 0.08 \\
\midrule
HydraLoRA & 82.93 & 84.02 & 71.05 & 87.69 & 46.25 & 74.39 & 0.28 \\
\midrule
A-LoRA-K & 83.34 & 85.18 & 72.33 & 88.15 & 46.85 & \textbf{75.17} & \textbf{0.08} \\
\bottomrule
\end{tabular}
\end{table*}

\subsection{Performance on highly dissimilar tasks}
\label{disimilar}
This is a crucial question about the limits of our alignment hypothesis. To address this, we ran a supplementary experiment using the exact experimental setup from Section~\ref{sec: increasing rank} / Table~\ref{qwen-flan}.

As detailed in ~\ref{Experiment 2: Increasing rank data}, this dataset (a subset of FlanV2) is specifically designed for high-task diversity, comprising \textbf{10 distinct task clusters}.  These tasks are highly dissimilar, including not only NLU (e.g., ANLI) and QA (e.g., ARC) but also \textbf{Struct-to-Text} (e.g., E2ENLG) and Translation.

We can confirm that \textbf{Align-LoRA still demonstrates a significant performance improvement} over the standard high-rank LoRA baseline in this highly dissimilar, multi-domain setting. This provides strong evidence that our alignment strategy is beneficial even when tasks have little in common, as it enhances the learning of foundational shared knowledge.

\begin{table*}[h!]
\centering
\caption{Performance comparison on highly dissimilar tasks}
\label{tab:dissimilar_tasks}
\begin{tabular}{l | c c c c c c}
\toprule
\textbf{Metrics} & \textbf{Base} & \textbf{LoRA} & \textbf{HydraLoRA} & \textbf{R-LoRA} & \textbf{M-LoRA} & \textbf{A-LoRA-K} \\
\midrule
7B & 39.82 & 48.18 & 49.12 & 49.51 & 49.74 & \textbf{50.09} \\
\midrule
14B & 45.33 & 52.74 & 53.76 & 54.08 & 54.18 & \textbf{54.47} \\
\midrule
\% Param & - & 0.22 & 0.25 & 0.25 & 0.22 & 0.22 \\
\bottomrule
\end{tabular}
\end{table*}

\subsection{Alignment on Multi-Head Architectures}
\label{Alignment on Multi-Head Architectures}
Align-LoRA represents a necessary "course correction" in the field: we demonstrate that for many scenarios, the complex "task-specific" modeling is effectively redundant. Our method achieves superior performance with \textbf{zero inference latency}, offering a distinct perspective from the trend of increasing complexity. We think that a future hybrid approach (Align first, then Decompose) is promising. We conducted a new experiment integrating our alignment into the R-LoRA and M-LoRA architecture. As shown below, adding alignment yields further gains even in multi-head structures.

\begin{table*}[h!]
    \centering
    \caption{Performance Comparison of Alignment Loss Integrated with Multi-Head Architectures (R-LoRA and M-LoRA).}
    \label{tab:alignment_on_multihead}
    \begin{tabular}{l | c c c c c c c c}
        \toprule
        \textbf{Model} & \textbf{Task1} & \textbf{Task2} & \textbf{Task3} & \textbf{Task4} & \textbf{Task5} & \textbf{Task6} & \textbf{Task7} & \textbf{Task8} \\
        \midrule
        R-LoRA & 89.80 & 62.51 & 89.36 & 83.78 & 95.12 & 91.02 & 92.17 & 50.15 \\
        M-LoRA & 91.35 & 62.51 & 91.98 & 84.70 & 95.93 & 91.02 & 91.97 & 50.20 \\
        R-LoRA+Align & 90.15 & 62.45 & 91.68 & 84.35 & 95.12 & 92.17 & 92.56 & 50.31 \\
        A-LoRA-K & \textbf{92.23} & 64.85 & \textbf{92.89} & 85.73 & 95.93 & 93.35 & 92.93 & 53.66 \\
        M-LoRA+Align & 92.15 & \textbf{64.93} & 92.15 & \textbf{85.95} & \textbf{96.32} & \textbf{93.63} & \textbf{93.27} & \textbf{53.90} \\
        \bottomrule
    \end{tabular}
\end{table*}

\section{Datasets}
\label{dataset}
This paper presents a compelling re-evaluation of LoRA-based multi-task learning paradigms, with a series of rigorous designs that strongly support its core claims.
The setup spans mainstream model families such as LLaMA and Qwen, with parameter scales ranging from 3B to 14B that is sufficiently large for the field of parameter-efficient fine-tuning, and it covers three distinct categories of task datasets. The ablation studies, including the analysis of head similarity, rank scaling effects, and cross-module validation, further validate the rationality of the proposed hypothesis about task-shared representations.
\subsection{Experiment 1}
\label{Experiment 1: observation}
In Section~\ref{observation}, we fine-tune Qwen2.5-3B on five tasks:
\begin{enumerate}
    \item \textbf{Natural Language Inference}: QNLI~\citep{wang2018glue}
    \item \textbf{Physical Question Answering}: PiQA~\citep{bisk2020piqa}
    \item \textbf{Word Relation Reasoning}: Winogrande~\citep{sakaguchi2021winogrande}
    \item \textbf{Closed-Book Question Answering}: ARC~\citep{allenai:arc}
    \item \textbf{Mathematical Reasoning}: GSM8K~\citep{cobbe2021training}
\end{enumerate}

\subsection{Experiment 2}
\label{Experiment 2: Increasing rank data}
In Section~\ref{sec: increasing rank}, following the setting of~\cite{tian2024hydralora, huang2024lorahub}, we utilize a subset of the \texttt{Flanv2} dataset~\citep{wei2022finetuned} for complex mixed multi-task and multi-domain scenarios. This dataset covers both Natural Language Understanding (NLU) and Natural Language Generation (NLG) tasks, which are grouped into 10 distinct task clusters. The specific datasets used in our experiments are sourced from LoRAHub (https://huggingface.co/datasets/lorahub/flanv2), a curated and open-sourced repository that provides organized access to these resources. Then we evaluate it with the Big-Bench Hard (BBH) benchmark~\citep{suzgun-etal-2023-challenging}, using accuracy as the core metric—higher accuracy directly reflects stronger model performance

We summarize the details of the used datasets as follows:
\begin{description}[style=unboxed, leftmargin=0pt, itemsep=3pt, parsep=0pt]
    \item[\textbf{Natural Language Inference}] 
    Infers logical relationships (entailment, contradiction, neutral) between sentence pairs. 
    \textit{Datasets:} ANLI, CB, MNLI, QNLI, SNLI, WNLI, RTE.
    
    \item[\textbf{Coreference Resolution}] 
    Identifies textual mentions referring to the same entity for contextual understanding. 
    \textit{Datasets:} DPR, WSC273.
    
    \item[\textbf{Struct-to-Text Conversion}] 
    Generates natural language descriptions from structured data inputs. 
    \textit{Datasets:} CommonGen, DART, E2ENLG, WebNLG.
    
    \item[\textbf{Closed-Book QA}] 
    Answers general knowledge questions without external information access. 
    \textit{Datasets:} ARC, NQ, TriviaQA.
    
    \item[\textbf{Sentiment Analysis}] 
    Determines sentiment polarity (positive/negative) of given text. 
    \textit{Datasets:} IMDB, Sentiment140, SST-2, Yelp.
    
    \item[\textbf{Commonsense RC}] 
    Integrates reading comprehension with reasoning beyond explicit content. 
    \textit{Datasets:} CosmosQA, ReCoRD.
    
    \item[\textbf{Paraphrase Detection}] 
    Judges whether two sentences convey equivalent semantic meanings. 
    \textit{Datasets:} MRPC, QQP, Paws Wiki.
    
    \item[\textbf{Translation}] 
    Converts text between languages while preserving meaning and subtleties. 
    \textit{Datasets:} En-Fr (WMT'14); En-De/Tr/Ru/Fi/Ro (WMT'16); En-Es (Paracrawl).
    
    \item[\textbf{Commonsense Reasoning}] 
    Applies physical, scientific principles and common sense in reasoning. 
    \textit{Datasets:} COPA, HellaSwag, PiQA, StoryCloze.
    
    \item[\textbf{Reading Comprehension}] 
    Derives answers from provided texts containing relevant information. 
    \textit{Datasets:} BoolQ, DROP, MultiRC, OBQA, SQuAD v1/v2.
\end{description}

\subsection{Experiment 3}
\label{Experiment 3 data}
First, to measure the ability to generalize, models are fine-tuned on the five-task benchmark introduced in Appendix~\ref{Experiment 1: observation} (QNLI, PiQA, Winogrande, ARC, and GSM8K). Performance is then evaluated on the challenging Big-Bench Hard (BBH) benchmark, which consists of complex reasoning tasks that were not seen during training. We adopt accuracy as the key metric here, with higher accuracy indicating better generalization of the model to these unseen complex reasoning tasks. This setup tests how well knowledge from the training domains is transferred to a different, more difficult set of problems.

Second, to assess the model's ability to adapt to the training domains, we conduct experiments on a broader eight-task benchmark. In this setup, models are evaluated on the corresponding test sets for each of the eight training tasks, with accuracy as the core metric. This evaluation measures the model's in-domain performance, and higher accuracy directly indicates stronger adaptive capability to the training domains. The eight tasks, categorized by their primary reasoning skill, are as follows:
\begin{enumerate}
    \item \textbf{Reading Comprehension}: BoolQ~\citep{clark2019boolq}
    \item \textbf{Science Question Answering}: SiQA~\citep{sap2019social}
    \item \textbf{Physical Commonsense}: PiQA~\citep{bisk2020piqa}
    \item \textbf{Word Relation Reasoning}: Winogrande~\citep{sakaguchi2021winogrande}
    \item \textbf{Commonsense Reasoning}: Hellaswag~\citep{zellers2019hellaswag}
    \item \textbf{Open-Book Question Answering}: OBQA~\citep{OpenBookQA2018}
    \item \textbf{Closed-Book Question Answering}: ARC (easy \& challenge)~\citep{allenai:arc}
    \item \textbf{Mathematical Reasoning}: GSM8K~\citep{cobbe2021training}
\end{enumerate}

Finally, for hyperparameter analysis, the model was trained on a 5-task dataset with the same settings as in the Appendix~\ref{Experiment 1: observation}, and the average performance is reported.

\section{Feature Visualization}
\label{sec: Feature Visualization}

\begin{figure}[h]
    \centering
    \includegraphics[width=0.48\textwidth]{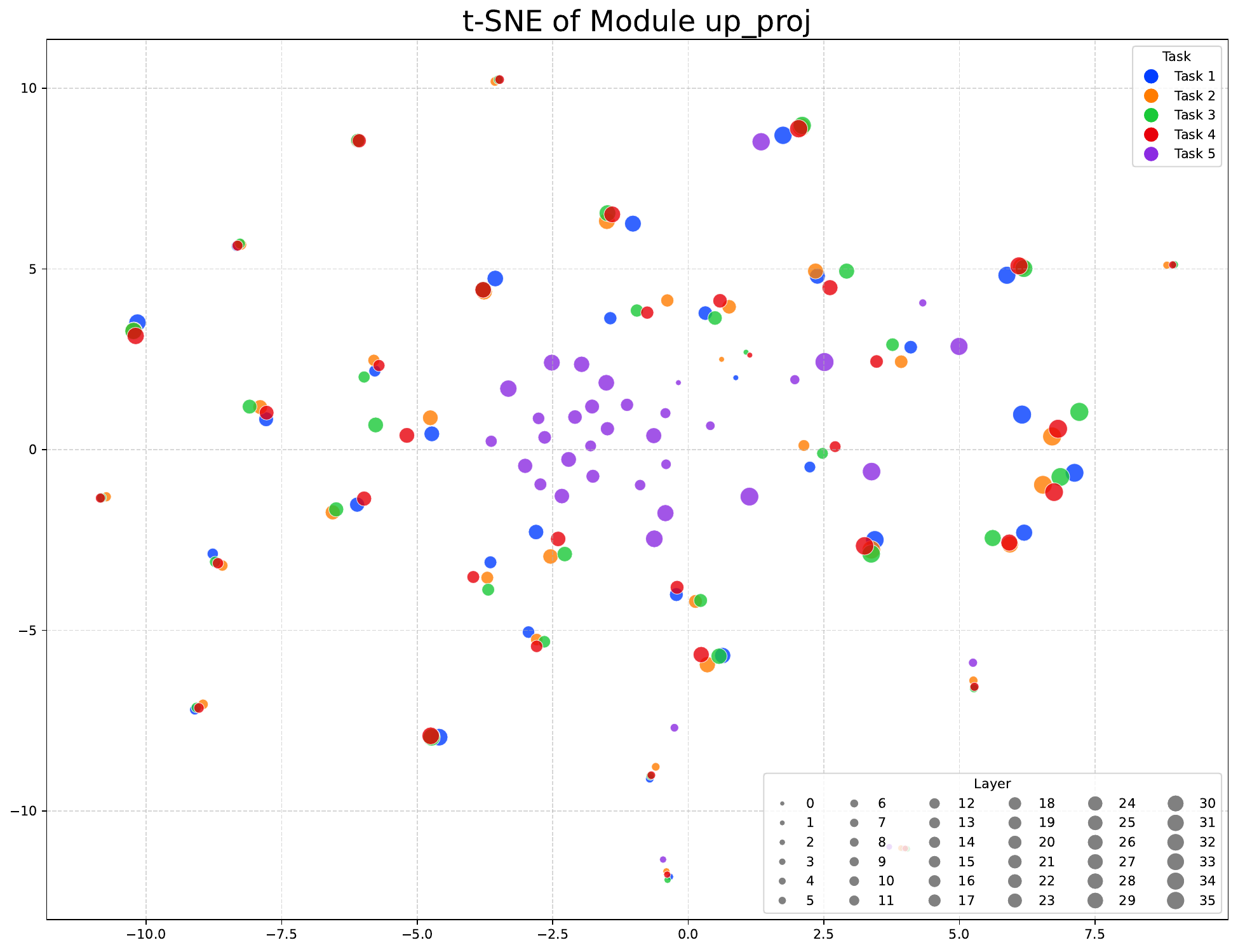}
    \hfill
    \includegraphics[width=0.48\textwidth]{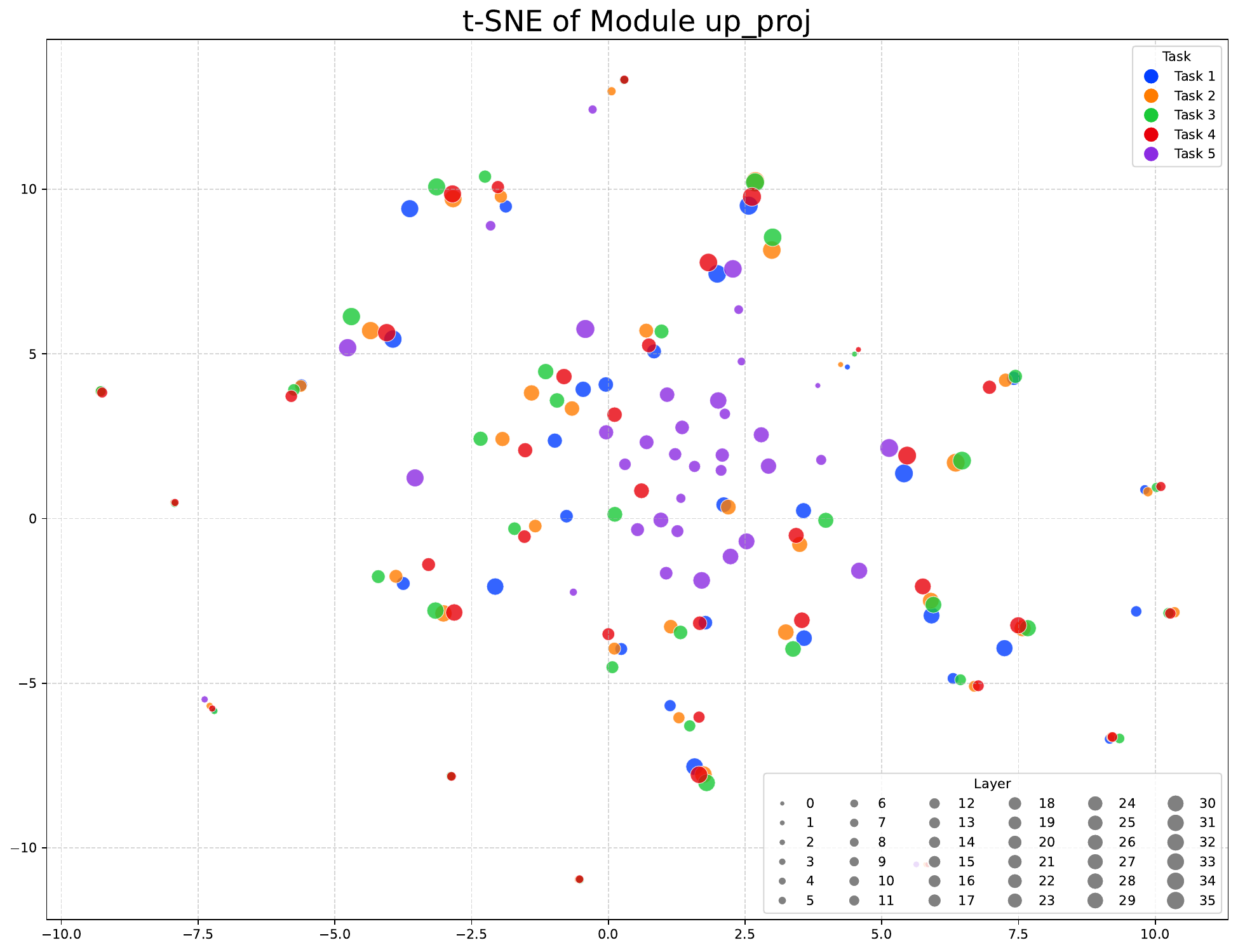}
    
    \caption{Qualitative t-SNE visualization of task representations for LoRA and Align-LoRA. Align-LoRA brings task representations closer in the shared latent space. Together with the intra-task control in Table~\ref{tab:inter-intra-alignment}, this result supports the interpretation that cross-task alignment contributes beyond generic same-task regularization.}
    \label{fig:feature-visualization} 
\end{figure}

We performed t-SNE analysis~\citep{maaten2008visualizing} on the representations of different tasks and compared LoRA with Align-LoRA, as shown in Figure~\ref{fig:feature-visualization}. Through representation alignment, Align-LoRA brings the task representations closer to each other, which helps the model learn general knowledge across tasks.

However, the goal is not to make the representations identical. Different tasks inherently contain distinct knowledge, so their representations must maintain a degree of separation to preserve task-specific information. Forcing the representations to become too close can lead to an "over-alignment" problem. This is consistent with our hyperparameter analysis in Figure~\ref{hyperparameter}, which demonstrates that an excessively large $\lambda$ value degrades performance by enforcing an overly aggressive alignment.

\section{Baseline}
\label{baseline}
\begin{enumerate}
    \item \textbf{LoraHub} randomly aggregates 20 LoRAs for new downstream tasks, employing a black-box optimization technique to determine each LoRA's weight without gradient computation of the large model (i.e., parameter-level weighted averaging).

    \item \textbf{LoRA MoE} orchestrates $n$ experts $E_1,\dots,E_n$ (each $E_i=B_iA_i$) with a router network $R$ parameterized by $W_R \in \mathbb{R}^{d_m \times n}$. A softmax over the Top-$K$ entries of $W_R^T x$ yields gating scores
\begin{equation}
    s_i = R(x)_i = \text{softmax}(\mathrm{Top}(W_R^T x, K)),
\end{equation}
and the output aggregates the gated expert outputs:
\begin{equation}
    y = \sum_{i=1}^n s_i \cdot E_i(x).
\end{equation}

    \item \textbf{HydraLoRA} uses a shared matrix $\mathbf{A}$ to project the input $x$ into a lower-dimensional space, with separate matrices $B_1,\dots,B_n$ modulating each expert's output:
\begin{equation}
    y = \sum_{i=1}^n s_i \cdot (B_i \mathbf{A} x).
\end{equation}

    \item \textbf{R-LoRA} extends HydraLoRA with multi-head randomization: it retains the shared $\mathbf{A}$ but independently initializes each $B_i$ and applies Dropout to the projected input to encourage head diversity:
\begin{equation}
    y = \sum_{i=1}^n g_i \cdot (B_i \cdot \mathrm{Dropout}(\mathbf{A} x)),
\end{equation}
where $g_i$ is the gating weight of the $i$-th head.
\end{enumerate}

\section{Implementation Details}
\label{implementation}
All models are trained on NVIDIA 4090 GPUs with bfloat16 mixed precision, a learning rate of $2\times10^{-4}$, \texttt{lora\_alpha}=32, dropout 0.1, warmup ratio 0.03, and a cosine annealing scheduler. Following HydraLoRA~\citep{tian2024hydralora}, the experiments in Table~\ref{llama-flan} adapt only \texttt{q\_proj} and \texttt{v\_proj} with rank 8, except the rightmost \texttt{LoRA}$^\dagger$, which uses rank 30 to match the parameter count of the multi-head variants; all other experiments adapt \texttt{down\_proj}, \texttt{up\_proj}, and \texttt{gate\_proj}. The number of heads equals the number of tasks, i.e., 5 for Table~\ref{table: observation} and 10 for Table~\ref{llama-flan}. The alignment weight is set to $\lambda=0.1$ for Align-LoRA-K and $\lambda=0.15$ for Align-LoRA-M.



\end{document}